\documentclass{article}
\usepackage{arxiv}
\usepackage{amsmath}
\usepackage{geometry}
\usepackage{layouts}
\usepackage{svg}
\usepackage{txfonts}
\usepackage[utf8]{inputenc} % allow utf-8 input
\usepackage[T1]{fontenc}    % use 8-bit T1 fonts
\usepackage{hyperref}       % hyperlinks
\usepackage{url}            % simple URL typesetting
\usepackage{booktabs}       % professional-quality tables
\usepackage{amsfonts}       % blackboard math symbols
\usepackage{nicefrac}       % compact symbols for 1/2, etc.
\usepackage{microtype}      % microtypography
\usepackage{lipsum}
\usepackage{graphicx}
\graphicspath{ {./images/} }

\title{Stochastic stability analysis of legged locomotion using unscented transformation}

\author{
Güner Dilsad Er \\
  Max Planck Institute for Intelligent Systems, Tübingen, Germany\\
 Middle East Technical University (METU), Ankara, Turkey\\
  \texttt{guenerdilsad.er@tuebingen.mpg.de} \\
   \And
 Mustafa Mert Ankaralı \\METU Robotics and AI Technologies Application and Research Center (METU-ROMER),\\
  Department of Electrical and Electronics Engineering, \\ Middle East Technical University (METU), Ankara, Turkey\\
  \texttt{mertan@metu.edu.tr}
}

\begin{document}
\maketitle
\begin{abstract}
 In this manuscript, we present a novel method for estimating the stochastic stability characteristics of metastable legged systems using the unscented transformation. Prior methods for stability analysis in such systems often required high-dimensional state space discretization and a broad set of initial conditions, resulting in significant computational complexity. Our approach aims to alleviate this issue by reducing the dimensionality of the system and utilizing the unscented transformation to estimate the output distribution. This technique allows us to account for multiple sources of uncertainty and high-dimensional system dynamics, while leveraging prior knowledge of noise statistics to inform the selection of initial conditions for experiments. As a result, our method enables the efficient assessment of controller performance and analysis of parametric dependencies with fewer experiments. To demonstrate the efficacy of our proposed method, we apply it to the analysis of a one-dimensional hopper and an underactuated bipedal walking simulation with a hybrid zero dynamics controller.

 %Keywords

  %   Stochastic stability
  %  Metastable legged systems
  %  Unscented transformation
  %  State space discretization
   % Initial conditions
   % Computational complexity
   % Dimensionality reduction
   % Output distribution estimation
   % Noise statistics
   % Uncertainty
   % High-dimensional system dynamics
   % Controller performance
   % Parametric dependencies
   % One-dimensional hopper
   % Underactuated bipedal walking simulation
   % Hybrid zero dynamics controller
\end{abstract}

% keywords can be removed

\section{Introduction}

Analytical models do not represent the real systems perfectly, because some phenomena like impact and friction always bring discrepancies to the actual implementation. In exchange for their agility and capabilities, legged robots are much more vulnerable to the stochastic effects of unknown terrains than other mobile robotic systems. They eventually encounter different abnormalities, such as slight differences in elevations (e.g., holes, rocks) or some contaminated surfaces that drastically affect friction and impact dynamics. Accordingly, controllers should take account of the external noises.  

Rhythmic gaits used for legged locomotion include walking, running, galloping, trotting, and pronking. Stance dynamics of those gaits are related to the restricted three-body problem and do not admit to a closed-form solutions \cite{SaranliHumanoid,poincareChaos}. 
Poincar\'e's return map analysis is frequently used to simplify the periodic trajectories' limit cycles \cite{Grizzle2001}. 
This simplification manages the stability characteristics deterministically and oversees the stochastic impacts of the external disturbances. 
Deterministic limit cycle stability analyses, like the eigenvalue analysis based on apex to apex numerical Jacobians \cite{Stride,Er2022}, are frequently utilized in literature to cover the stability characteristics of legged systems. 
However, those deterministic numerical methods are frequently difficult/expensive to apply, requiring a calculation around each fixed point.
Furthermore, since systems no longer have actual limit cycles under persistent disturbances, they are theoretically incorrect in the presence of stochastic disturbance. For instance, the MARLO robot was able to walk in a lab setting, but it only managed a few steps before falling while testing outside due to a slight incline in the walkway \cite{Griffin2017}. To increase stability, it is necessary to take into account the characterization of walking's stochastic dynamics during system identification and controller design. In the robotics community, this is a significant but understudied method that is primarily handled as a robust control problem \cite{Dai2012, Griffin2017} rather than an examination of stochastic dynamics.

Legged locomotion is characterized by the dynamic interactions between the feet and the contact surface. The term dynamic locomotion usually stands for an unbalanced walking cycle leading to a stable gait behavior. Underactuated legged systems leverage the underactuation to achieve dynamic locomotion. As the trade-off between stability and agility in the control theory, there exists an essential relationship between stability and maneuverability for legged systems \cite{Aoi2016}. Under disturbance, as in many stochastic dynamical systems, legged robots exhibit long-living, locally stable behaviors up to some point that cannot handle the external effects anymore. Once the disturbed system's states go into a region with a different attractor, the system behavior irreversibly adjusts to the new local dynamics. In simpler words, a legged robot can run for quite some time, but it will definitely fall due to the external stochastic effects. Since they eventually leave the locally stable gait behavior, they cannot be considered ``stable'' (at least asymptotically). On the other hand, they obviously operate for long periods of time, making calling them ``unstable'' would be mistaken. A need arises for a new type of classification in the control theory. In line with several studies in the locomotion community \cite{Byl2009,Byl2008a,Ankarali2014} , we believe that Metastability is a well-defined candidate for defining this phenomenon.

The legged counterpart of the metastable systems substitutes the metastable equilibrium with dynamic locomotion and the absolute minimum with falling to the ground. Hereby, walking is well-characterized as a metastable process. In the literature, Byl and Tedrake introduced the conceptual connection and utilized the metastability concept to quantify the stochastic stability of rimless wheel and compass-gait walking on rough terrain \cite{Byl2008thesis, Byl2009}. They also utilized stochastic optimization to improve the overall stability of legged systems. 

Byl and Tedrake's metastable limit cycle analysis methodology deals with walking systems by their closed-loop return map dynamics. Following their main methodology, we represented the return maps as Markov chains for the first step. States of this Markov chain consist of mesh points (and their vicinity) of the state space for the particular legged system. Then we obtained state transition matrices of this Markov chain from Monte Carlo Simulations by integrating the system simulations from each mesh point for different values of terrain slope. Monte Carlo analysis is a common method to estimate a probability distribution's progress over time. The method is based on selecting a sufficient number of representative random samples and simulating them with the dynamical system model. The result is the probability density as an estimate of the system output. 
This experimental future state-value distribution can be visualized as a histogram and used for assessing the estimation error. As the number of mesh points, which has been taken as initial conditions to simulations, increases, the accuracy of the future state-value distribution will increase. 

Referring to Byl's studies \cite{Byl2008a}, Benallegue and Laumond questioned the computational feasibility of the prior method for complex walking systems and proposed a solution for the legged systems with high-dimensional states using the limit-cycle property of stable walking \cite{Benallegue2013}. Actually, the biggest problem in the former approach is computational complexity which comes from two different aspects: simulation time and meshing methodology. Firstly, obtaining the state transition matrix requires numerous simulations to get the most accurate results. For instance, for a 5-link bipedal walking simulation, a one-step simulation takes up to 0.5 seconds in MATLAB. Running $10^6$ simulations lasts more than five days, which can be reduced to less than one day by parallel programming, but it will still be too long to run enough experiments to build a smooth stochastic return map. In addition to the infeasibility of conducting thousands of experiments, if experimental setups are in the loop, another problem arises; physical damage. The more experiment we conducted, the more likely the system will fall due to its metastable nature. Therefore, the system is more likely to get damaged. Secondly, for lower dimensional (1-DOF or 2-DOF) systems with one-dimensional noise, meshing the state space will be quite easy as slicing a range of values or meshing a surface. However, as the number of dimensions increases, meshing a cube or a 4D structure becomes more complex, even impossible. For example, for a 5-link bipedal robot, one must discretize all the state-space in ten dimensions along with the noise space. If the noise comes from only one source, noise space discretization is relatively trivial. Whereas, in the case of multiple noise sources, the former method fails to present an efficient way to analyze the dynamics. For 3D walking, the required degree of freedom increases quickly. Subsequently, Saglam and Byl introduced an improved meshing technique \cite{Saglam2014b}, but unfortunately, this improvement did not break the curse of dimensionality. On the other hand, Saglam and Byl's studies contributed to the previous breakthrough to handle legged systems as metastable systems by compilation of the methodology \cite{Saglam2014a}, new meshing technique \cite{Saglam2015a} and optimal controller designs \cite{Saglam2018}. Now, we propose a new methodology to further improve metastable analysis by borrowing estimation methods from stochastic tools.

Kalman filtering \cite{Kalman1960} is a well-known estimation method widely used in robotic platforms. 
The Extended Kalman Filter (EKF) extends the classic Kalman Filter for nonlinear systems where nonlinearity is approximated using the first or second-order derivative. It tries to capture nonlinearity using Taylor expansion around a local point. 
On the other hand, the nonlinear extensions of the Kalman Filter consist of nonlinear propagation of probability densities. The sample-and-propagate methods can be generalized as perturbation methods, using samples as initial values, which are perturbations from the mean trajectory. Sampling transforms the continuous state domain into a discrete set of points. EKF fails to conduct nonlinear propagation because of its basis for linearization and partial derivatives instead of propagation \cite{Grewal2015}. 

Unscented Kalman Filter is a special case of sigma point filters introduced to improve filtering performance. Unscented transformation \cite{Julier2004, Wan2006} is a powerful tool to estimate the statistics of a random variable that undergoes a nonlinear transformation \cite{Safaoui2021} and is used in many applications ranging from sensor fusion for state estimation \cite{Choi2021} to an unscented Kalman observer \cite{Daid2021}. Moreover, in recent studies, Sieberg et al. combined an artificial neural network with confidence level adjustment and presented a hybrid state estimation structure using unscented transformation \cite{Sieberg2021}. We borrowed this useful stochastic tool to make informed choices on initial conditions for the stochastic analysis simulations. Eliminating computational complexity, we can utilize the mean first passage time metric to characterize the stochastic stability of high-dimensional underactuated nonlinear systems. Additionally, unlike the previous studies \cite{Byl2009, Saglam2014b}, estimation with unscented transformation allows us to deal with multiple sources of uncertainties on higher dimensional systems. 

Even though unscented transformation helps estimate the future state-value distribution for nonlinear systems, nonlinear transformation does not help with the higher-order moments. The higher-order moments of the estimation distributions are not tracked. Many possible distributions share the same mean and variances having distinct higher-order moments. Sample-and-propagate methods can capture the exact and unique solution if the transformation results are linear. In the existence of nonlinearities, we cannot reach the exact solution. Because there is no unique solution, estimation performance assessments for this type of estimator are also tricky. Therefore, tuning the parameters of the filters to reach a better estimation can be done by comparing the estimation with the results of the Monte Carlo experiments. This tuning procedure gives the proper parameters only for this particular nonlinear system. Tuning the weights in the unscented transformation-based estimation affects the estimated results, so one should be careful when choosing those parameters. This paper assumes that the future state-value distribution is a Gaussian, so the future state-value distribution is built as a Gaussian with estimated mean and variances. We compared the estimated mean and variances with the results from Monte Carlo experiments.

This study proposes a more efficient estimation method for metastable system properties based on unscented transformation; therefore, there will be no need for conducting many experiments through Monte Carlo sampling. We implemented our method to examine the stochastic stability of a one-dimensional hopper and an idealized 5-link biped simulation with a hybrid zero dynamics controller under disturbance. The one-dimensional hopper states an example of a simple-legged system. After observing the satisfactory estimation results, we extended the methodology to a higher dimensional system. This 5-link walker model is inclusive for robot walkers due to its nonlinear, underactuated, and hybrid nature.

\section{Methodology}

\subsection{System as an Absorbing Markov Chain}

Markov chains are stochastic models that describe a sequence of possible events whose probability only depends on the previous event \cite{Gagniuc2017}. One way of performing metastable limit cycle analysis in stochastic rhythmic dynamical systems is by representing the discrete-time system dynamics as a Markov process. 

In nature, we observe that behaviors including walking, running and many other types of legged locomotion are periodic. The system dynamics defined on a periodic return map allows us to discretize the system behavior by Poincare maps. To build Poincare maps, we preferred apex point during the gait, where the vertical velocity of the robot is zero. 

We represented the apex to apex dynamics as a discrete system for the stochastic stability analysis. 
\begin{equation}  
\begin{aligned}
\mathbf{x}_{k+1}=\mathbf{f}(\mathbf{x}_{k},\mathbf{w}_{k})
\end{aligned}
\end{equation}
where $\mathbf{x}_{k}$ and $\mathbf{w}_{k}$ represent the states and noises at time step $k$ respectively. We assumed the noise values to be drawn from a Gaussian distribution with zero mean and covariance of $R_\omega$.

The stochastic state transition dynamics can also be modeled with an infinite Markov chain; however, in general, approximated as a finite-state Markov process via discretization of the state-space into a finite set of states \cite{Byl2009}. State discretization allows us to compute and analyze a finite-state Markov chain model of the system. The state space is divided into N pieces and assigned to Markov states. The state transition matrix $\mathbf{T}_{N\times N}$ of this Markov chain collects the transition probabilities between the $N$ predefined states. The probability of transition from state $i$ to state $j$ is,
\begin{equation}
\begin{aligned}
    \mathbf{T}_{ij}=\mathbb{P}(\mathbf{x}_{k+1}=s_j|\mathbf{x}_{k}=s_i)\label{nonabsorbingTranProb}.
\end{aligned}
\end{equation}
An absorbing Markov chain is a Markov chain with at least one state that is ``impossible'' to leave. 
In legged locomotion, we can consider the absorbing state collecting all configurations where the robot falls \cite{Byl2009}, or we can simply assign that state variables associated with some unwanted behaviors to the absorbing state, assuming that recovery from these is impossible.
Besides, we can specify a particular region we would like to operate and take the other configurations that belong to the absorbing state. Assuming $s_1$ is the absorbing state, we can state the following,
\begin{equation}
    \mathbf{T}_{11} = 1 \quad \text{and} \quad \mathbf{T}_{1j}=0 \quad \text{for} \quad j \neq 1.
\end{equation}
Absorbing Markov chains has one eigenvalue at $\lambda_1=1$ and the stable distribution matrix of that Markov chain will be the first eigenvector of $\mathbf{T}$ in \eqref{stateTran} which is the first unit vector, which means this system will eventually stop at the first (absorbing) state. The second-largest magnitude eigenvalue of the matrix $\mathbf{T}$ corresponds to the largest magnitude eigenvalue of the $\Bar{\mathbf{T}}$ and is related to the metastable characteristic of the system. The eigenvector associated with the largest magnitude eigenvalue of $\Bar{\mathbf{T}}$ describes the long-living (metastable) distribution of the state.
\begin{equation}
    \mathbf{T}=\begin{bmatrix}\mathbf{1}_{1\times 1}&0_{1\times N-1}\\\mathbf{T}_{j1_{N-1\times 1}}&\Bar{\mathbf{T}}_{N-1\times N-1} \end{bmatrix}\label{stateTran}
\end{equation}
The state transition matrix of the Markov chain can be built via Monte Carlo simulations. According to the law of large numbers, the average of the results obtained from a large number of trials should be close to the expected value and tends to become closer to the expected value as more trials are performed \cite{Dunn2012}. However, building the state transition matrix with Monte Carlo simulations requires too many trials (simulations or experiments), which is highly inconvenient for complex legged systems. Thus, we propose a method based on Unscented transformation and prefer to choose sigma points using prior knowledge on noise characteristics and simulate the system accordingly. As a result, the unscented transformation concept results in a more efficient estimation of state transition matrices.

\subsection{State-Value Distribution Estimation}

\subsubsection{Linearization Based Estimation}

The core component in metastability analysis that we adopt in this study is computing/deriving the probability of the following probability density function, $p(\mathbf{x}_{k+1} | x_{k})$, which is technically the process update step in Kalman based state-estimators. One way of approximating this step is adopting the same principle with Extended Kalman Filter, where we can formulate a covariance prediction either using a linearized analytical model of the system or a numerically linearized version. Unfortunately, almost all valid-legged locomotion models do not allow analytic dynamics integration; thus, we are limited to numerical linearization of the return map dynamics. 

We formulated the linearization-based estimation technique for a generalized case of nonadditive noise. This formulation begins with the nonlinear system description in Markov chains.
In this paper, We apply the noise at the impact instant, and its effect on the future state-value state $\mathbf{x}_{k+1}$ is not explicitly known (and nonlinear).
\begin{equation}
\begin{aligned}
\mathbf{x}_{k+1}=\mathbf{f}(\mathbf{x}_{k},\mathbf{w}_{k}), \quad \mathbf{w}_k \sim \mathcal{N}(0,\mathbf{Q}_k)
\end{aligned}\label{eqn:ekfsys}\end{equation}
At the model forecast step \eqref{ekf_forecast}, the forecast value of $\mathbf{x}_{k+1}$ (indicated by $\mathbf{x}_{k+1}^{f}$) is produced by propagating the initial optimal estimate $\mathbf{x}_{k}^a$ through the nonlinear system and used to compute the mean and covariance of the forecast value of $\mathbf{x}_{k+1}$. In this thesis, the initial optimal estimate is defined as the initial condition, more clearly, the midpoint of the range defining each Markov state.
\begin{equation}
    \begin{aligned}
        \mathbf{x}_{k}^a:=\mathbf{x}_{k}
    \end{aligned}
\end{equation}
The predictable part of $\mathbf{x}_k$, i.e., the forecast value, is given by
\begin{equation}
\begin{aligned}
\mathbf{x}_{k+1}^{f}=&\mathbb{E}[\mathbf{x}_{k}]=\mathbb{E}[\mathbf{f}(\mathbf{x}_{k},\mathbf{w}_{k})]
    \end{aligned}\label{forecast_first}
\end{equation}
The next step is expanding system description $\mathbf{f}(.)$ in Taylor Series about the optimal estimate $\mathbf{x}^a$ as follows,
\begin{equation}
    \begin{aligned}
        \mathbf{f}(\mathbf{x}_{k},\mathbf{w}_{k})\approx \mathbf{f}(\mathbf{x}_{k}^a,0)+\mathbf{f}_{\mathbf{x}_k}(\mathbf{x}_{k}^a,0)(\mathbf{x}_k-\mathbf{x}_{k}^a)+\mathbf{f}_{\mathbf{w}_k}(\mathbf{x}_{k}^a,0)\mathbf{w}_k + H.O.T.
    \end{aligned}\label{ekf_taylor}
\end{equation}
where $\mathbf{f}_{\mathbf{x}_k}=\frac{\partial \mathbf{f}}{\partial \mathbf{x}_k}$, $\mathbf{f}_{\mathbf{w}_k}=\frac{\partial \mathbf{f}}{\partial \mathbf{w}_k}$. Higher order terms are ignored and the partial derivatives with respect to state and noise should be calculated by numerical methods. Then forecast value is calculated by substituting \eqref{ekf_taylor} into \eqref{forecast_first},
\begin{equation}
    \begin{aligned}
        \mathbf{x}_{k+1}^f&\approx\mathbb{E}\left[\mathbf{f}(\mathbf{x}_{k}^a,0)+\mathbf{f}_{\mathbf{x}_k}(\mathbf{x}_{k}^a,0)\underbrace{(\mathbf{x}_k-\mathbf{x}_{k}^a)}_{e_k}+\mathbf{f}_{\mathbf{w}_k}(\mathbf{x}_{k}^a,0)\mathbf{w}_k\right]\\
          \mathbf{x}_{k+1}^f&\approx\mathbf{f}(\mathbf{x}_{k}^a,0)+\mathbf{f}_{\mathbf{x}_k}(\mathbf{x}_{k}^a,0)\underbrace{\mathbb{E}[e_k]}_{0}+\mathbf{f}_{\mathbf{w}_k}(\mathbf{x}_{k}^a,0)\underbrace{\mathbb{E}[\mathbf{w}_k]}_{0}\\
         \mathbf{x}_{k+1}^f&\approx\mathbf{f}(\mathbf{x}_{k}^a,0)
    \end{aligned}
\end{equation}
The forecast error equation becomes as follows:
\begin{equation}
\begin{aligned}
e_{k+1}^f&=\mathbf{x}_{k+1}-\mathbf{x}_{k+1}^f\\
&=\mathbf{f}(\mathbf{x}_{k},\mathbf{w}_{k})-\mathbf{f}(\mathbf{x}_{k}^a,0)\\
&\approx\mathbf{f}_{\mathbf{x}_k}(\mathbf{x}_{k}^a,0)e_k+\mathbf{f}_{\mathbf{w}_k}(\mathbf{x}_{k}^a,0)\mathbf{w}_k
\end{aligned}\label{ekf_forecast}\end{equation}
The forecast error covariance is calculated as
\begin{equation}
    \begin{aligned}
        \mathbf{P}_{k+1}^{f}&=\mathbb{E}[e_{k+1}^f(e_{k+1}^f)^T]\\
        &=\mathbb{E}\left[ (\mathbf{f}_{\mathbf{x}_k}(\mathbf{x}_{k}^a,0)e_k+\mathbf{f}_{\mathbf{w}_k}(\mathbf{x}_{k}^a,0)\mathbf{w}_k)(\mathbf{f}_{\mathbf{x}_k}(\mathbf{x}_{k}^a,0)e_k+\mathbf{f}_{\mathbf{w}_k}(\mathbf{x}_{k}^a,0)\mathbf{w}_k)^T\right]\\
        &=\mathbf{f}_{\mathbf{x}_k}(\mathbf{x}_{k}^a,0)\mathbb{E}[e_k(e_k)^T](\mathbf{f}_{\mathbf{x}_k}(\mathbf{x}_{k}^a,0))^T+\mathbf{f}_{\mathbf{w}_k}(\mathbf{x}_{k}^a,0)\mathbb{E}[\mathbf{w}_k(\mathbf{w}_k)^T] (\mathbf{f}_{\mathbf{w}_k}(\mathbf{x}_{k}^a,0))^T\\
        &=\mathbf{f}_{\mathbf{x}_k}(\mathbf{x}_{k}^a,0)\mathbf{P}_{k}(\mathbf{f}_{\mathbf{x}_k}(\mathbf{x}_{k}^a,0))^T+\mathbf{f}_{\mathbf{w}_k}(\mathbf{x}_{k}^a,0)\mathbf{Q}_{k} (\mathbf{f}_{\mathbf{w}_k}(\mathbf{x}_{k}^a,0))^T\\
    \end{aligned}
\end{equation}
The estimation in this study considers a one-step calculation of successive states. One can reformulate the future state-value mean and variance equations for the one-step calculation instead of a recursive calculation of forecast and error values with time points such as $k$ and $k+1$. Since the initial value for $\mathbf{x}_k$ ($=\mathbf{x}_0$) is deterministically known, $\mathbf{P}_0$ will be 0, and estimated mean and variances for the next step can be computed as
\begin{equation}
\begin{aligned}
\boldsymbol{\mu}_{1} &= \mathbf{x}_{1}^{f}=\mathbf{f}(\mathbf{x}_{0}^a,0), \\
\mathbf{P}_{1}^{f} &= \mathbf{f}_{\mathbf{w}}(\mathbf{x}_{0}^a,0)\mathbf{Q}_{0} \left(\mathbf{f}_{\mathbf{w}}(\mathbf{x}_{0}^a,0)\right)^T
\end{aligned}\label{mean_var_ekf}\end{equation}
Using the estimated future state-value mean $\boldsymbol{\mu}_{1}$ and future state-value variance $\mathbf{P}_{1}^{f}$, the estimated normal distribution can be constructed as $\mathbf{X}_{1} \sim \mathcal{N}\left(\mu_{1}, \mathbf{P}_{1}^{f}\right)$. 

\subsubsection{Unscented Transformation Based Estimation}

The fundamental motivation behind using unscented transform is that approximating a probability distribution is easier than approximating an arbitrary nonlinear function \cite{Julier2004}. Instead of approximating the system equations by linearization, we calculated sigma points and used them in the unscented transformation to directly approximate the output probability density functions. The assumption under the probability distribution estimation is to have a Gaussian noise and expect the output distributions to be Gaussian. The central limit theorem states that the sampling distribution approaches a normal distribution as the sample size increases \cite{Fischer2011}. That is why it can be assumed that, under the exposure of multiple noise sources, the future state-value distributions will coincide with a Gaussian distribution. 

The formulation steps are similar to the Unscented Kalman Filters \cite{Julier2004, Wan2006}. First of all, formulation for the generalized case of nonadditive noise requires an augmented state definition $\mathbf{x}_k^{a}$ with system states $\mathbf{x}_k$ and zero mean noises $\mathbf{w}_{k}$,
\begin{equation}  
\mathbf{x}_k^{a}=\begin{bmatrix}\mathbf{x}_k^T & \mathbf{w}_{k}^T\end{bmatrix}^T\label{augmented}.\end{equation}
Previously known nonlinear system dynamics $\mathbf{f}$ and the noise variance characteristics $\mathbf{P}_{k}$ are,  
\begin{equation}  
\begin{aligned}
\mathbf{x}_{k+1}=\mathbf{f}(\mathbf{x}_{k}^{a}), \quad
\mathbf{P}_{k}=\begin{bmatrix}\varepsilon&0\\0&R_w\end{bmatrix},
\end{aligned}
\end{equation}
where $\mathbf{P}_{k}$ contains the known variances as diagonal entries, and $R_w$ represents the noise variances. Since our initial states $\mathbf{x}_{k}$ are deterministic, their variance will be zero; however, for computational purposes, we specified their variance as a very small value $\varepsilon$. If we take the state variance as zero, that will cause a problem in the matrix square root step. 

Sigma points represent the chosen initial conditions so that the output of the nonlinear system to these initial conditions will provide the information related to output distribution. The sigma point set $\mathbf{X}_{k}$ \eqref{sigmaset} contains $2n+1$ sigma points $\mathbf{x}_{k}^j$ and their associated weights $\mathbf{W}^j$ so that their mean will be $\mathbf{x}_{k}^a$ and variance $\mathbf{P}_{k}$, where $n$ is the dimension of augmented state.
\begin{equation}
\begin{aligned}
    \mathbf{X}_{k}&=\{ ( \mathbf{x}_{k}^j , \mathbf{W}^j) | \quad & j=0\dots 2n   \}\\
    \mathbf{x}_{k}^0&=\mathbf{x}_{k}^{a}, \quad& -1<\mathbf{W}^0<1 \\
      \mathbf{x}_{k}^j&=\mathbf{x}_{k}^{a} + A_j, \quad& j= 1 \dots n\\
          \mathbf{x}_{k}^j&=\mathbf{x}_{k}^{a} - A_j, \quad& j= n+1 \dots 2n\\\mathbf{W}^j&=\frac{1-\mathbf{W}^0}{2n}, \quad& j= 0 \dots 2n\\
          A_i&=\left(\sqrt{\frac{n}{1-\mathbf{W}^0}\mathbf{P}_{k}}\right)_i
\end{aligned}\label{sigmaset}
\end{equation}
The weight of the first sigma point $\mathbf{W}^0$, in \eqref{sigmaset}, controls the proximity of sigma points to their mean. If $\mathbf{W}^0 \leq 0$ or $\mathbf{W}^0 > 0$, the sigma points tend to be closer or further from the origin.  

At the model forecast step \eqref{forecast}, the transformed points ($\mathbf{x}_{k+1}^{f,j}$) are produced by propagating each sigma point through the nonlinear system and used to compute the mean and covariance of the forecast value of $\mathbf{x}_{k+1}$.
\begin{equation}
    \mathbf{x}_{k+1}^{f,j}=\mathbf{f}(\mathbf{x}_{k}^{f,j})\quad j= 0 \dots 2n \label{forecast}
\end{equation}
After forecasting, estimated mean and variances are computed as
\begin{equation}
\begin{aligned}
    \boldsymbol{\mu}_k&=\sum^{2n}_{j=0} \mathbf{W}^i_m \mathbf{x}_{k+1}^{f,j},\\
    \mathbf{P}^f_{k+1}&=\sum^{2n}_{j=0} \mathbf{W}^i \{ \mathbf{x}_{k+1}^{f,j} -\boldsymbol{\mu}_k \} \{ \mathbf{x}_{k+1}^{f,j} -\boldsymbol{\mu}_k \}^T.
    \end{aligned}
\end{equation}
Using the estimated output mean $\boldsymbol{\mu}_0$ and output variance $\mathbf{P}^f_{1}$, we can construct the estimated normal distribution ${\mathbf{X}_{output}\sim\mathcal{N}(\mathbf{\mu}_0,\mathbf{P}^f_{1})}$.

\subsection{Analysis and Performance Metric}

The core purpose of this paper is to assess the system's stability in the existence of noise. Investigating the state transition matrix, we can infer the stochastic characteristics and comment on the effect of noise for each configuration. We calculated the transition probabilities of nonabsorbing states in \eqref{nonabsorbingTranProb} as the following,
\begin{equation}
\begin{aligned}
    \mathbf{T}_{ij}=&\mathbb{P}(\mathbf{x}_{k+1}=s_j|\mathbf{x}_{k}=s_i)\\=&\mathbf{F}_{\mathbf{X}_{output}}(\frac{s_{j+1}+s_j}{2})-\mathbf{F}_{\mathbf{X}_{output}}(\frac{s_{j}+s_{j-1}}{2})
    \end{aligned}
\end{equation}
where $\mathbf{F}_{\mathbf{X}_{output}}$ represents the cumulative distribution function of output distribution.
Transition probabilities to the absorbing state are equal to the total probability of not going into nonabsorbing states.
\begin{equation}
\begin{aligned}
      \mathbf{T}_{i1}&=\mathbb{P}(\mathbf{x}_{k+1}= s_1|\mathbf{x}_{k}=s_i)\\&=1-\sum_{i=0}^N \mathbb{P}(\mathbf{x}_{k+1}\neq s_1|\mathbf{x}_{k}=s_i)
      \end{aligned}
\end{equation}
Finally, setting transition probabilities from the absorbing state to zero, we complete the state transition matrix structure in \eqref{stateTran}.

Mean first passage time is a stability metric for metastable systems and extracted from the second-largest magnitude eigenvalue of the state transition matrix $\mathbf{T}$. Definition of system-wide mean first passage time value $M$ is the following;
\begin{equation}
    M=\frac{1}{1-\lambda_2}.\label{mfpt}
\end{equation}
We can assess the performances of different controllers by the closed-loop system's mean first passage time values. In addition, we can compare the closed-loop system behaviors with respect to different levels of noise variances.

State-dependent mean first passage time curves are one of the properties to discuss. The state dependent MFPT vector $m$ collects the expected passage time from the state $s_i$ to the absorbing state $s_1$. This vector is computed as in \eqref{state_dep_mfpt}.
\begin{equation}
    \begin{aligned}
        m=\begin{bmatrix} 0 \\ (\mathbf{I}-\mathbf{\Bar{T}})^{-1}\mathbf{1} \end{bmatrix}
    \end{aligned}\label{state_dep_mfpt}
\end{equation}
where $\mathbf{\Bar{T}}$ is the state transition matrix without its first row and first column, $\mathbf{I}$ is the identity, and $\mathbf{1}$ is a vector with all elements equal to 1.

Finally, there are some metastable properties worth to discuss. Metastable neighborhood is actually the stochastic counterpart of the fixed point of the deterministic return map. This neighborhood indicates the joint probability of the two successive body angular velocity value measured just before the impact. And, metastable distribution is the stationary distribution of the Markov states unless the robot is not in the absorbing state and calculated by replacing the first element of the eigenvector associated with the second largest magnitude eigenvalue with zero and normalizing it. Then the metastable neighborhood, i.e., joint probability, can be calculated by multiplying the state transition matrix with the metastable distribution.

\section{Example: Analysis of One Dimensional Hopper}

This hopper with one-dimensional vertical motion is expected to demonstrate the applicability of the proposed stochastic analysis methodology. Its quasi-linear nature prohibits generalizing the deductions on more complex dynamical systems; however, promising results are introduced in the following parts.

\begin{figure}[htb]
    \centering
    \includegraphics[scale=0.7]{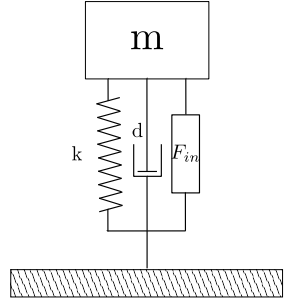}
    \caption{Illustration of the hopper with one-dimensional vertical motion}
    \label{fig:oneleg}
\end{figure}

Chosen one-dimensional hopper model is a variation of the Spring Loaded Inverted Pendulum (SLIP) template with constant forcing and damping called F-SLIP recently studied by \cite{Tanfener2022}. Using SLIP model variations brings the advantage of simplicity for implementation and analysis together with its applicability to many legged systems as a template \cite{Saranli2003}. Figure~\ref{fig:oneleg} depicts the dynamical model of the F-SLIP template model with one-dimensional vertical motion.

\begin{figure}[htb]
    \centering
    \includegraphics[width=\textwidth]{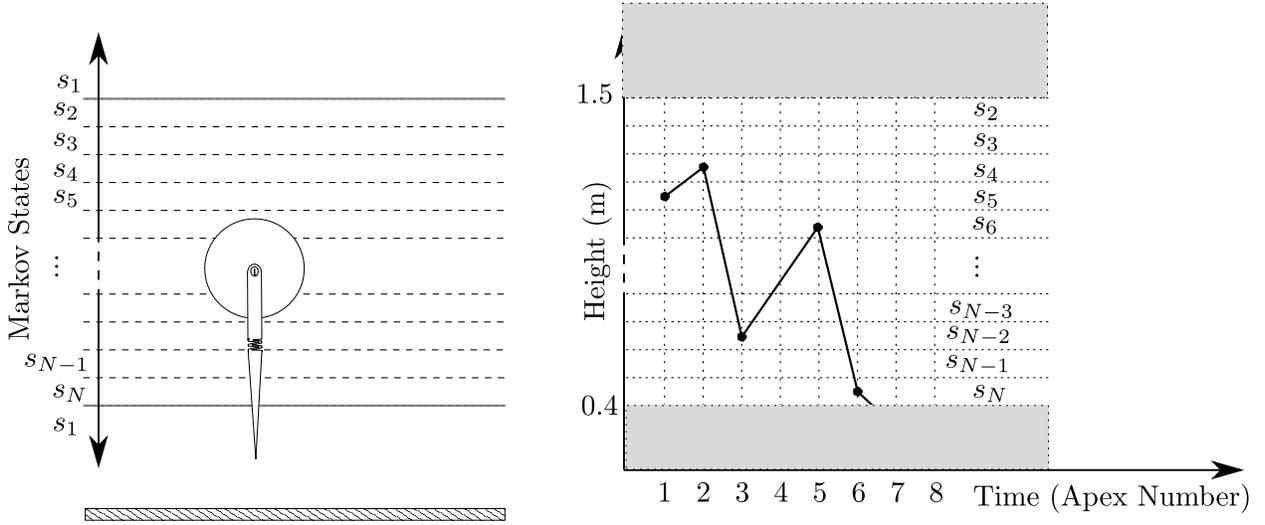}
    \caption{(on the left) Discretization of states for one dimensional hopper. The apex height of the top of of the leg is discretized into a finite set of slices. (on the right) An example passage time observation. Passage time is observed as 7, that means the robot leaves the predetermined region at the $7^{th}$ step.}
    \label{fig:one_leg_slice}
\end{figure}

For the stochastic analysis, the first step is representing the system as an absorbing Markov chain. Since this system has a one-dimensional state representation, determining the Markov states is pretty straightforward. States of the Markov chain are obtained by discretizing the state space, in this case, only the height values, using equally spaced 220 slices between $0.4$ and $1.5$ m and defining an absorbing state to represent the height values below $0.4$ and higher than $1.5$, this slicing is roughly illustrated in Figure~\ref{fig:one_leg_slice}. 

This paper covers three main approaches to estimating the state transition matrix. First of all, one can calculate the state transition matrix by running the aforementioned systematic experiments covering a wide range of noise values. This method can be interchangeably addressed as Monte Carlo simulations when the slicing of noise values is infeasible, so the initial conditions are randomly sampled in space. Monte Carlo simulation results are expected to be very close to systematic experiments if enough number of experiments are conducted. Secondly, one can use a linearized version of the system to calculate the mean and variance of the future state-value distribution. This linearization can be conducted either, if available, using the analytically linearized version of the system or numerically calculated linearized system matrices (Jacobians) at respective points. The third approach is proposed to handle nonlinear systems more efficiently without losing the information sourced from nonlinearity. For the cases where nonlinear behavior is dominant, the estimation method based on unscented transformation is expected to outperform the linearization-based methods because unscented transformation is supposed to handle nonlinearities and takes the actual system dynamics into account for calculations.

\begin{table}[!htb]
\begin{center} \caption{Comparison of different transition matrix estimation methods}
 \begin{tabular}{|c||c|c|c|c|c|}
 \hline
 Method &\# of experiment & $\lambda_1$ & $\lambda_2$ &  $\lambda_3$ &  $\lambda_4$\\
 \hline \hline
Systematic Experiments    & $6.6 \cdot 10^6$ &1.0000&    0.9917&    0.8602&    0.7328\\
Monte Carlo &       $3.3 \cdot 10^5$ &      1.0000&    0.9916&    0.8595&    0.7343 \\
Unscented Transform & $6.6 \cdot 10^2$ &     1.0000&    0.9877&    0.8534&    0.7252  \\
Linearization &$4.4 \cdot 10^2$&  1.0000  &  0.9877  &  0.8534  &  0.7252\\
\hline
\end{tabular}
 \label{tab:BMUcomp}  
\end{center}
\end{table}

In order to compare different estimation methods, we can examine the state transition matrices' eigenvalues that contain essential information about the Markov chain and subsequently the metastable system itself. From the second largest magnitude eigenvalue, system-wide mean first passage time is calculated to use as an indicator for stochastic stability. Table~\ref{tab:BMUcomp} shows the first four eigenvalues of the state transition matrices for impact velocity noise variance of 0.05 for comparison. The methods based on unscented transformation and linearization give almost the same result and slightly deviated results from Monte Carlo and systematic experiments. The linearization-based estimation method is observed to give almost identical results to the proposed method.

\begin{figure}[htb]
    \centering
    \includegraphics[scale=0.7]{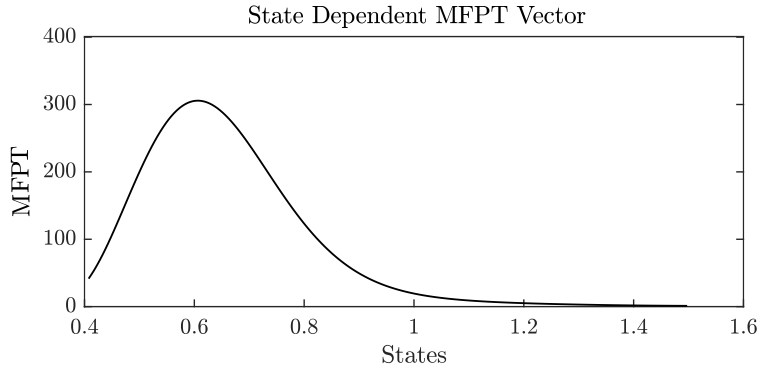}
    \caption{State dependent MFPTs, quantifies the relative stability of each point in state space, for impact velocity noise variance of 0.05.}
    \label{fig:stateMFPT}
\end{figure}

State-dependent MFPT curve is plotted in Figure~\ref{fig:stateMFPT}, using the state transition matrix estimated by the proposed method. Each initial condition has a particular state-dependent MFPT $m(s_i)$ and $m(s_i)$ quantifies the relative stability for each point. Different from the rimless wheel (RW) in \cite{Byl2009}, for this system, state-dependent MFPT curve is far from flat. Therefore, the objective system can be inferred as highly sensitive to initial conditions. In addition, the same conclusion can be reached by investigating the eigenvalues: $\lambda_1=1$, $\lambda_2=0.9917$, $\lambda_3= 0.8602$, $ \lambda_4=0.7328$. The value of $\lambda_3$ means that almost 14\% of the contribution to the probability function at the initial condition is lost ("forgotten") with each successive step. Again, this was not the case for the rimless wheel in \cite{Byl2009}. RW system has its third eigenvector near 0.5 and forgets 50\% of the initial condition. As a result, within a few steps, initial conditions for any wheel beginning in the range of analysis have therefore predominantly evolved into the metastable future state-value distribution unless it fails. Analogously, the motion of the one-dimensional hopper will converge to its metastable distribution after more steps but eventually it will.

\section{Example: Analysis of Bipedal Walking}

\begin{figure}[htb]
    \centering
    \vspace{2mm}
    \includegraphics[width=0.35\textwidth]{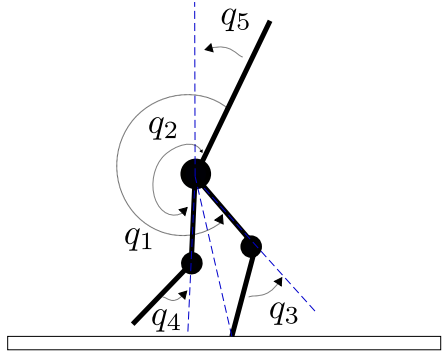}
    \caption{Illustration of 5-link bipedal robot}
    \label{fig:biped}
\end{figure}

We will showcase our method on a 5-link bipedal walking model, the RABBIT \cite{rabbitRef}. The simulation testbed has a controller design based on optimization of the hybrid zero dynamics (HZD) following the same steps in \cite[Chapter 6.6.2.1]{HZDBook}. As in the implementation in \cite{sovukluk_2022}, we defined the system dynamics as 
\begin{equation}
    \dot{x}=f(x)+g(x)u,\label{dyn}
\end{equation}
where ten dimensional state $x:=[q^T \enspace \Dot{q}^T]^T$ collects the configuration variables $q:=[q_1\enspace q_2\enspace q_3\enspace q_4\enspace q_5]^T$, as shown in Figure~\ref{fig:biped}, along with their velocities.

HZD ensures that relative angles $h_0(q)$ track desired trajectory $h_d(q)$. Definition of tracking error (y) is
\begin{equation}
    y=h(q):=h_0(q)-h_d(q) \label{hzd}.
\end{equation}
Control input applied by the HZD controller takes the following form
\begin{equation}
    u=(\mathcal{L}_g\mathcal{L}_fh)^{-1}(-\mathcal{L}^2_fh+\varv),\label{input}
\end{equation}
where $\mathcal{L}_g\mathcal{L}_fh$ and $\mathcal{L}_f^2h$ represents Lie derivatives of tracking error with respect to system dynamics $f$ and $g$ in \eqref{dyn}. The control input is saturated in the implementation to make the simulation more realistic.

Study in \cite{HZDBook} proves that, with a simple PD controller, the solution of the closed-loop system converges to an exponentially stable periodic orbit of the hybrid zero dynamics. Therefore, we preferred to utilize a PD controller for $\varv$ to force $h$ in \eqref{hzd} to zero.
\begin{equation}
    \varv= K_{D} \mathcal{L}_{f} h+ K_{P} h
\end{equation}
Table \ref{tab1} shows different parameter choices for diagonal entries of $K_P$ and $K_D$ pairs to analyze the closed-loop behavior later with our proposed method.
\begin{table}[htb]
\caption{Different controller parameter pairs, values of the table represents the diagonal entries of $K_P$ and $K_D$ values}
\begin{center}
\begin{tabular}{|c||c|c||}
\hline
& $K_P$ & $K_D$\\
\hline\hline
\textbf{$C_1$ }& $[60 \enspace 90\enspace90 \enspace50]$ & $[10 \enspace 20\enspace20 \enspace10]$\\ \hline
\textbf{$C_2$ }& $[ 5\enspace 5\enspace 5\enspace5]$ & $[ 5\enspace 5\enspace 5\enspace5]$\\ \hline
\textbf{$C_3$ }& $[ 40\enspace 40\enspace 40 \enspace 40]$ &0 \\ \hline
\textbf{$C_4$ }& $[ 40\enspace 40\enspace 40 \enspace 40]$ & $[ 1\enspace 1\enspace 1\enspace1]$\\ \hline
\textbf{$C_5$ }& $[10 \enspace 89\enspace83\enspace50]$& $ [5.4 \enspace21\enspace 21\enspace 9]$ \\ \hline
\hline
\end{tabular}
\label{tab1}
\end{center}
\end{table}

The first step towards the stochastic analysis of bipedal locomotion is building the reachable state space. We need to find reachable state space by Monte Carlo sampling or meshing the space by predefined ranges \cite{Benallegue2013}. In the case of a 5-link bipedal robot, each Markov state $s_i$ can be chosen as a $10\times1$ vector, containing each link's angular positions and velocities. Nevertheless, it is too complicated to specify the Markov states for a 10D space. The underactuated 5-link bipedal testbed needs a different approach for stochastic analysis.

The system's controller follows a trajectory such that the unactuated link shows the desired behavior, i.e., actuated degree of freedoms indirectly control the body angle. As shown in Figure~\ref{fig:biped}, the position and velocity of stance and swing legs of the walker are defined relative to the body of the system. This coordinate configuration strengthens the idea that high bandwidth actuated joints are expected to be around their desired trajectory as long as the unactuated joint is close to its desired evolution. That is why observing body angle provides strong information about other joints' evolution and stability of the locomotion. Furthermore, for the reachable state space construction, the underactuated body angle $s_i=q_5^i$ and body angular velocity $s_i=q_{10}^i$ are suitable candidates to focus, search the vicinity of the fixed point and define the reachable limits assuming noises for all five states representing velocities. Nevertheless, this model reduction should be justified quantitatively. The next section explains the model reduction process for this 5-link bipedal system. 

\subsection{Extension to the High-Dimensional Systems} 

The main problem in extending this stochastic analysis to multidimensional systems stems from the requirement of meshing of multidimensional state spaces. In Saglam's studies \cite{Saglam2015a}, meshing the hybrid zero dynamics (HZD) surface is presented as an alternative to finding and meshing the reachable state space for a bipedal walker operated by an HZD controller. In this way, a switching mechanism between multiple HZD controllers becomes possible to increase stability. Despite the effort to decrease the complexity of the meshing process, the issue still exists and grows with the increasing degree of freedom and the variety of noise sources.

Linearization is a candidate method to decrease system order to identify and control systems. Numerical Jacobian calculations with variable step size can be conducted exactly in \cite{NumMATLAB}, or other methods such as analytical linearization, forward and center difference approximations can be used, noting that the linearization method will influence the result. After linearization, by investigating the eigenvector associated with the largest magnitude eigenvalue, the state can be identified such that the system is the most sensitive against a change in that state. Eigenvalues of the system give a picture of stability around the chosen operating point. The selected state can be used as the indicator state for the stability conditions in the stochastic stability analysis. Nevertheless, calculating the linearized system matrix with variable step size gets more difficult as the dimension increases. Choosing a fixed step size to tackle complexity diminishes the accuracy of the calculation.

Alternatively, this analysis for the most vulnerable state can also be done with stochastic tools. 
This study features PCA to reduce the objective dimensions to assess the legged system's stochastic stability. PCA is used as a preprocessing technique to reduce dimensionality. It aims to increase interpretability while minimizing information loss \cite{Jolliffe2016} and allows to use of previously collected data rather than conducting experiments for numerical simulation. A PCA plot converts the correlations among all cells into a 2D graph and allows to comment on the features in the dataset. The mathematical details of the method are not in the scope, so they are not covered in this paper. More detailed information can be referred from \cite{Jolliffe2016}.

The most critical limitation of PCA is its reliance on linear models and sensitivity towards outliers. Because PCA is a linear projection, it assumes a linear relationship between features and cannot capture the nonlinear dependencies. Its goal is to find the directions (i.e., principal components) that maximize the variance in a dataset.

We generalized the PCA to select "the most important state" in a multi-dimensional legged system. In this paper, we assumed to know the system's input-output relation. While having the system that generates the dataset, analyzing the dynamics based on just the data might seem controversial. However, this model reduction method can be generalized for a model reduction on experimental legged setups when this methodological study goes forward with the actual implementation.

\begin{figure}[htb]
     \centering
     \includegraphics[scale=0.7]{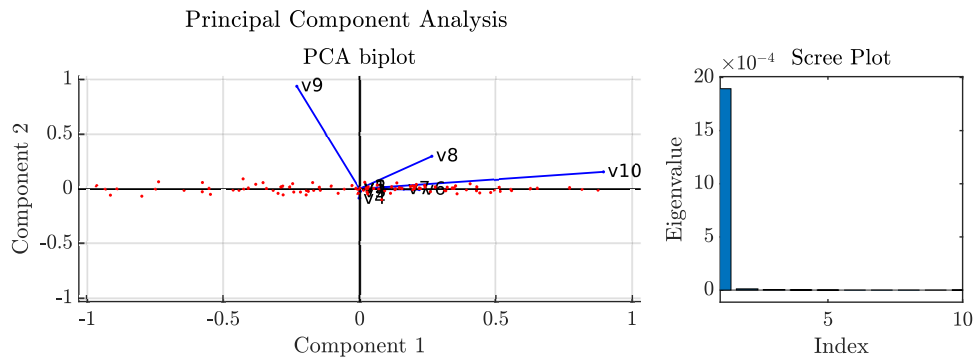}
     \caption{Principal Component Analysis for the 5-link bipedal system}
     \label{fig:biped_pca}
 \end{figure}

The bipedal walker is run for 100 steps, and a $10 \times 100$ dataset is generated. Next, using the built-in function for PCA in MATLAB, the dataset is visualized. Figure~\ref{fig:biped_pca} demonstrates the PCA biplot, scree plot, and score plot for the dataset. From the scree plot, it can be concluded that the first principle component (PC1) can be concluded as enough to describe the data. Investigating states' projections onto principal components, one can see that the $10^{th}$ state has the largest projection on the first principal component, meaning that one can roughly characterize the motion using the $10^{th}$ state.
Both linearization followed by eigenvector analysis and PCA give the same result about the $10^{th}$ state because they both rely on the linearity assumption for the state relations. And they both produced results that are compliant with our intuition. 

\subsection{Stochastic Analysis}

\begin{figure}[htb]
    \centering
     %trim={<left> <lower> <right> <upper>}
    \includegraphics[scale=0.9]{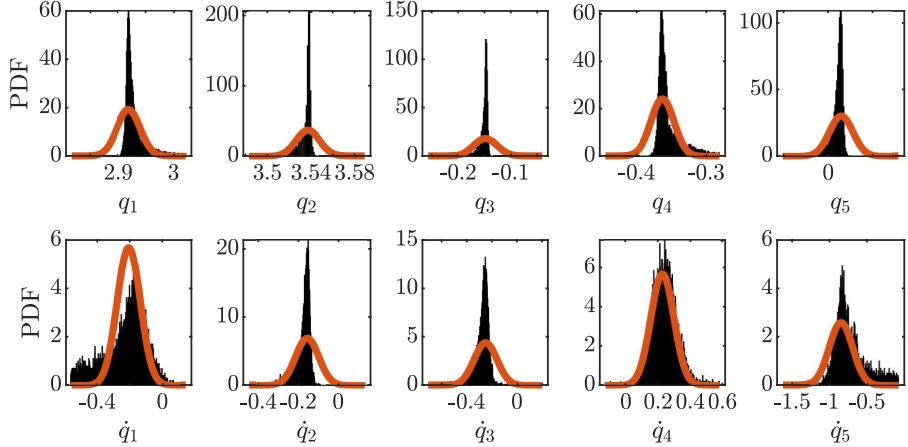}
    \caption{Comparison of experimental results and estimation with unscented transformation. Each subplot represents the future state-value distribution of the 5-link bipedal robot under disturbance. Histogram of experimental results is a product of $10^4$ experiments and our proposed method run 41 simulations for estimation.}
    \label{fig:Estimation}
\end{figure}

After choosing the Markov chain states, we need to build the state transition matrices by following the proposed methodology. Figure~\ref{fig:Estimation} shows that our experimental observation will support the claim that under Gaussian noise, future state-value distributions shapes, shown as histograms, also approach Gaussian shape, and the estimation is capable of capturing each ten state's variance and mean under multiple sources of noise. 

Building a state transition matrix with Monte Carlo simulations requires a repetitive calculation of many simulation (For Figure~\ref{fig:Estimation}, it is $10^4$ for one Markov state, $1.15\times10^6$ in total.) for the mid point of each state of the Markov chain as the initial condition of body angular velocity. Suppose we choose to build our matrices with Monte Carlo experiments. In that case, due to its random nature, no two matrices will be the same, and increasing the number of trials per Markov state, for example, from $10^4$ to $10^6$ leads to a lower variance among the produced matrices. In other words, Monte Carlo simulations result in different matrices for each experiment set requiring significant computational power and time. Apparently, Monte Carlo experiments are infeasible for high-dimensional cases, as also clearly stated in \cite{Benallegue2013}. Our method includes additional information to the experimental procedure to choose initial conditions to estimate each future state-value distribution. It requires $41$ experiments for each Markov state in a 5-link bipedal walking case.

\begin{figure}[htb]
    \centering
    \includegraphics[scale=0.7]{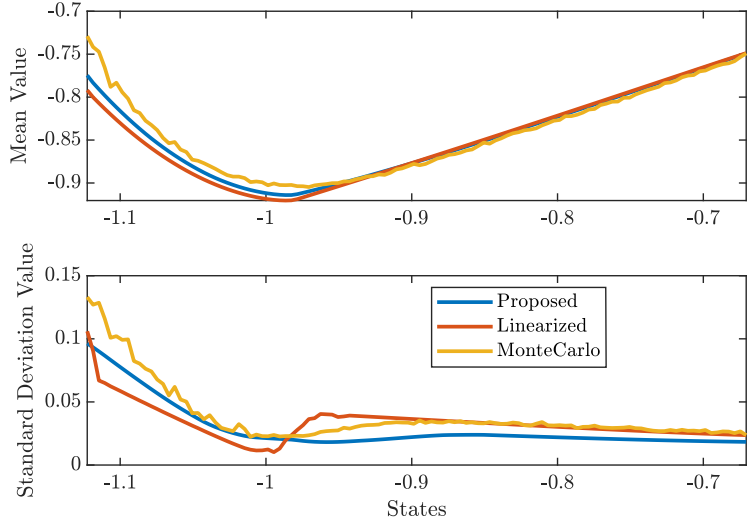}
    \caption{Comparison of mean and variances of future state-value PDF's }
    \label{fig:compmv}
\end{figure}

Mean and variances of estimated PDFs are plotted in Figure~\ref{fig:compmv} in order to assess the estimation performance. Around the region where the nonlinear behavior is dominant ($[-1.12,-0.95]rad/s$), the proposed method with unscented transformation works better than the linearization-based method as expected. In the linear region, both methods have a satisfactory performance; however, the linearization-based method works better. This may caused from the asymmetrical shape of the future state-value distribution, which can be observed in Figure~\ref{fig:Estimation}. The estimation can be improved by tuning the weights in the unscented transformation up to some level. After all, the aim is to find an approach with fewer assumptions and generalizable for highly nonlinear systems. Therefore, the unscented transformation is adopted for further investigation of the stochastic behavior of the bipedal system.

\begin{figure}[htb]
    \centering
    \includegraphics[scale=0.55]{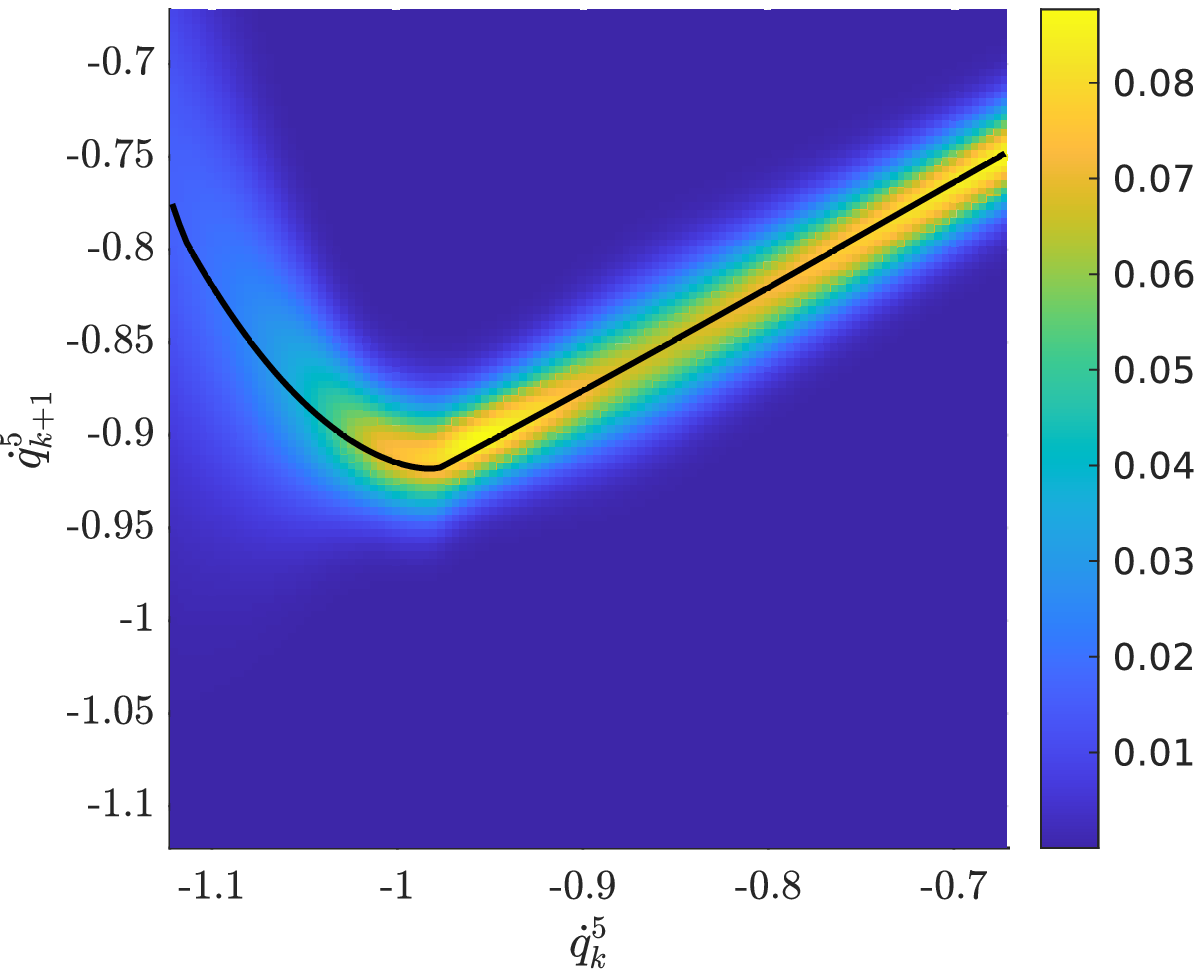}
    \includegraphics[scale=0.55]{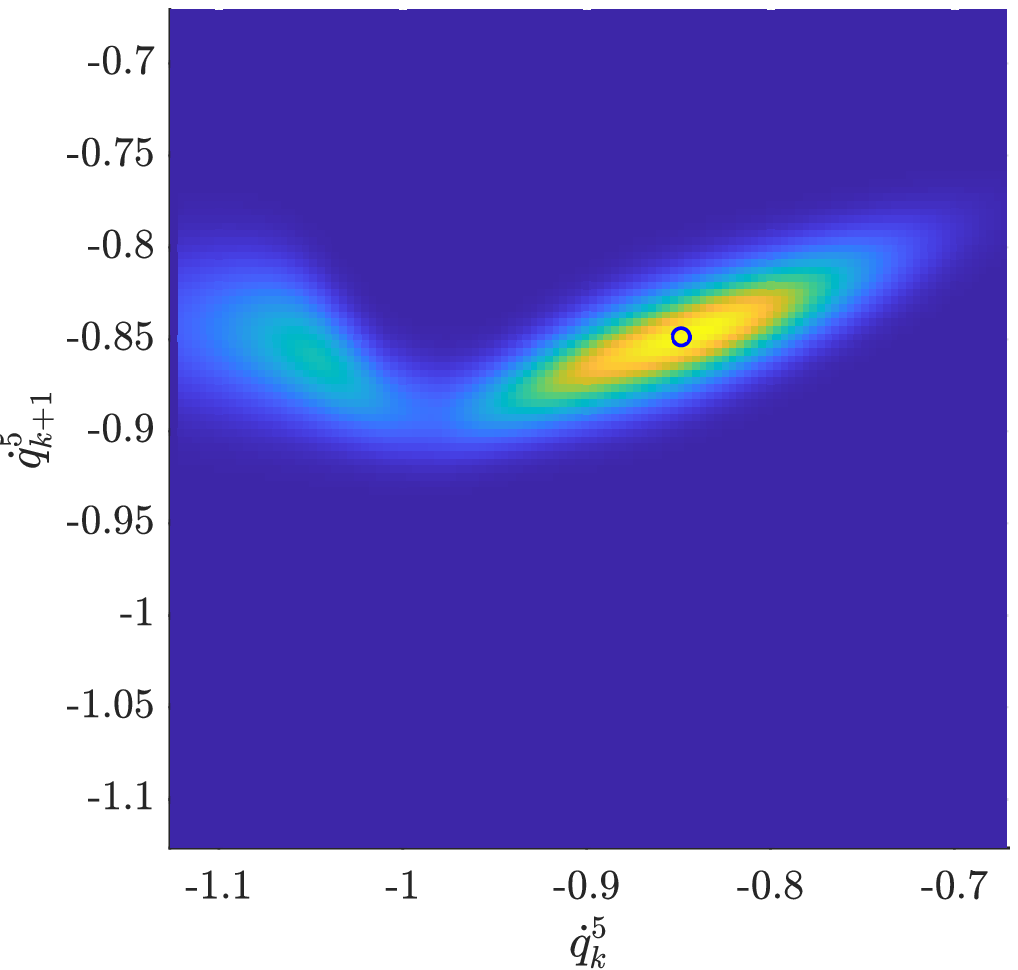}
    \caption{(on the left) State Transition Matrix for body angular velocity (i.e. stochastic returnmap) visualized together with deterministic return map of body angular velocity. 
    Colorful surface plot represents the $115\times115$ state transition matrix of body angle for a zero mean Gaussian noise with variance of $10^{-3}$ on each state. The black line represents the deterministic return map of body angular velocity. Controller is $C_1$ in Table \ref{tab1}. (on the right) Metastable neighborhood of state transitions for the bipedal system }
    \label{fig:UTstateTranMat}
\end{figure}

Figure~\ref{fig:UTstateTranMat} depicts the constructed state transition matrix for body angle together with a deterministic return map. We represented the future state-value distribution of body angular velocity if all the velocity states are subject to noise with known characteristics.

The state transition matrix represents a stochastic return map and includes valuable information about the system behavior. For example, For example, Figure~\ref{fig:UTstateTranMat} shows that the deterministic return map tends to be linear for some intervals, including the fixed point. That means, near the fixed point, the system can be assumed as a linear system. Under stochastic disturbance, estimation with the linearity assumption works very well, as seen in Figure~\ref{fig:compmv}. In addition, both deterministic and stochastic return maps indicate that the linearity assumption is not valid for a small region, so the body’s behavior cannot be generalized as linear. The stochastic return map implicates the same facts as the deterministic one and brings the future state-value variance information for different initial conditions of the body angular velocity. And the metastable neighborhood map in Figure~\ref{fig:UTstateTranMat}focuses on relating the probabilities of successive steps. This metastable neighborhood represents that if the robot starts to walk from an initial condition (whose attractor is this fixed point), its states at the next steps will most likely be around the fixed point unless it fails. 

In addition, we can infer our system's sensitivity to initial conditions by investigating the eigenvalues of the state transition matrix of the absorbing Markov chain: $\lambda_1=1$, $\lambda_2=0.9775$, $\lambda_3= 0.3552$, $ \lambda_4=0.2918$. The value of $\lambda_3$ means that nearly $65\%$ of the contribution to the probability function at the initial condition is lost ("forgotten") with each successive step. As noise variance increases, we observed that $\lambda_3$ decreases, which means the system tends to forget its initial condition more.

\begin{figure}[htb]
    \centering
  \includegraphics[scale=0.7]{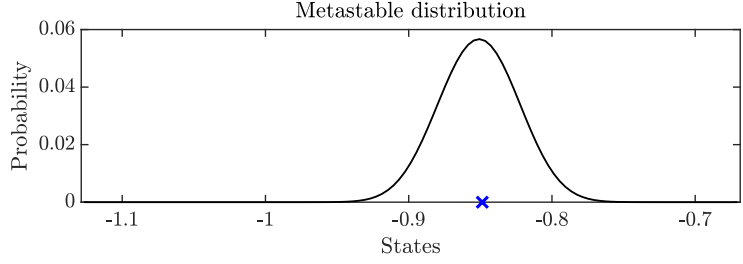}
    \caption{Metastable distribution for the bipedal system for noise variance of $10^{-3}$}
    \label{fig:biped_eigenvec}
\end{figure}

As stated before, the eigenvector associated with the second largest magnitude eigenvalue is used to calculate metastable distribution. Metastable distribution in Figure~\ref{fig:biped_eigenvec} states that if the body angular velocity starts from a random point, it will more likely be around $[-0.9, -0.8]rad/s$ unless it fails.

\begin{figure}[htb]
    \centering
    \includegraphics[scale=0.7]{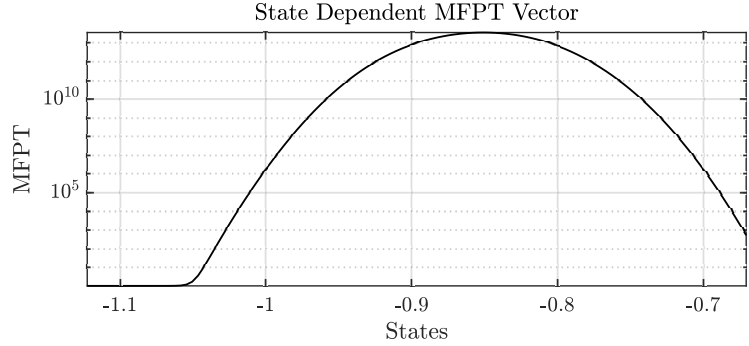}
    \caption{State dependent MFPTs for the bipedal system for noise variance of $10^{-3}$ }
    \label{fig:state_dep_mfpt}
\end{figure}

The state dependent MFPT vector in Figure~\ref{fig:state_dep_mfpt} also shows the initial conditions such that the system is more likely to maintain its locomotion under noise. The curves in Figures~\ref{fig:biped_eigenvec} and \ref{fig:state_dep_mfpt} imply the same possibilities.

\begin{figure}[htb]
    \centering
    \includegraphics[scale=0.7]{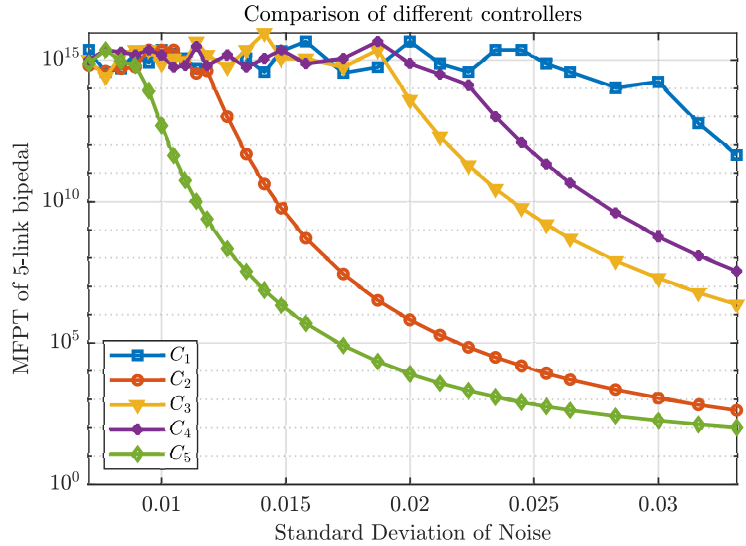}
    \caption{System-wide MFPT for the bipedal walker with respect to standard deviation of state noises, $\sigma$, obtained by unscented transformation method}
    \label{fig:MFPTcomparison}
\end{figure}

The feasibility of building the state transition matrix with unscented transformation allows us to assess the different controllers and analyze the system's stability under different noise levels. Each velocity state of the 5-link bipedal are subject to noise with the same variance. Figure~\ref{fig:MFPTcomparison} shows the dependence of system-wide MFPT in \eqref{mfpt} on noise standard deviation for different controller parameters in Table \ref{tab1}. The MFPT values over $10^{14}$ are not reliable due to MATLAB's numerical limits. From the figure, it can be deduced that the first controller $C_1$ is more stable than the other experimented controllers. Also, it can be related that the proportional controller $C_5$ shows the least stable behavior. Control input saturation prevents making this observation without conducting the stochastic analysis.

\section{Conclusion}

This study compiles the methodology for the stochastic analysis of legged systems. We proposed an efficient methodology for the stochastic stability analysis of metastable legged systems, allowing us to analyze system properties and compare the controllers' stability for various noise levels. 

The methodology is mainly based on the metastable nature of the legged locomotion. Under uncertainties, Metastable-legged robots can maintain their locomotion for a reasonable amount of time but eventually will fail. The idea of modeling the system as an absorbing Markov chain comes from this metastability. Stochastic return maps calculated as the state transition matrices for the absorbing Markov chains represent the system behavior in the existence of uncertainties. It is important to extract the information related to stochastic stability, such as metastable distribution and mean first passage time. In this study, those properties are investigated and used in commenting on the metastable behavior and controller comparison.

Inspiring from Kalman filters, we formulated two different estimation method to estimate state transition matrices and compared with the existing methods. At first, we formulated a linearization-based method, inspired by the extended Kalman filters, for the non-additive noise case. This method relies on the numerically linearized version of the system and fails for the systems with dominating nonlinear behavior. This limitation is an expected tradeoff originating from the linearization and we should eliminate it by extending the methodology for the dynamical systems.

In basic terms, suppose the objective system is a simple nonlinear system with additive noise inside. In that case, the effect of noise on the resulting apex-to-apex map is unknown because the noise input might go through a nonlinear transformation. This fact leads to the use of methods that take the nonlinear dynamics into account. In \cite{Ankarali2014}, the closed-loop system identification uses Poincar\'e theory and assumes the closed-loop juggling behavior as operating near a limit cycle. So, the authors fit a linear system model of apex-to-apex dynamics, assuming that the subject humans remain within a local region where the linear dynamics dominate. Therefore, an autoregressive Gaussian model using the apex height data can model the system. However, this approach cannot be generalized to all hybrid rhythmic dynamical systems, which may have some dominating nonlinear behavior and are not self-stable.

Unscented transformation provides a strong tool for estimation, requires much less simulation time to conduct, and leads to a low variance solution. Most importantly, it considers the full system dynamics and brings no simplification at this step while reducing the number of experiments. However, even if unscented transformation considers the whole system dynamics, some tradeoffs exist. For example, it assumes that the future state-value distribution is a Gaussian. As long as the future state-value is near symmetric, it does not cause a huge problem in capturing the mean and variances of the future state-value distribution. However, symmetry is non-generalizable to every system. In addition, another simplification step in the methodology related to the model reduction of the 5-link bipedal robot extends the stochastic stability analysis to multi-dimensional systems. The limitation comes from making dimension reduction based on methods that adopt linearization, such as Principal Component Analysis. This model reduction may require to be conducted with less assumption for highly nonlinear systems. In the end, despite these limitations of the principal component analysis and unscented transformation-based estimation, simulation results in this chapter showed that the improved stochastic stability analysis methodology provides a much faster analysis that is inclusive for complex legged systems. 

To conclude, this study discusses the preeminence of the proposed method based on unscented transformation over existing methods. Furthermore, we employed our estimation method for the stochastic stability analysis of legged systems with different specifications. Despite all the limitations, the proposed methodology satisfies the requirements of the analysis. Most importantly, this study promises a wide range of new research questions in this unexplored territory of legged locomotion.

%\bibliography{mendeley}
%\bibliographystyle{ieeetr}

\end{document}